\documentclass{ieeetj}
\usepackage{cite}
\usepackage{amsmath,amssymb,amsfonts}
\usepackage{algorithmic}
\usepackage{graphicx,color}
\usepackage{textcomp}
\usepackage{xcolor}
\usepackage{hyperref}
\hypersetup{hidelinks=true}
\usepackage{algorithm,algorithmic}
\usepackage{etoolbox}
\usepackage{placeins}
\BeforeBeginEnvironment{tabular}{\begingroup\footnotesize\setlength{\tabcolsep}{4pt}}
\AfterEndEnvironment{tabular}{\endgroup}
\def\BibTeX{{\rm B\kern-.05em{\sc i\kern-.025em b}\kern-.08em
    T\kern-.1667em\lower.7ex\hbox{E}\kern-.125emX}}
\AtBeginDocument{\definecolor{tmlcncolor}{cmyk}{0.93,0.59,0.15,0.02}\definecolor{NavyBlue}{RGB}{0,86,125}}

\def\OJlogo{\vspace{-4pt}$<$Society logo(s) and publication title will appear here.$>$}
\def\seclogo{\vspace{10pt}$<$Society logo(s) and publication title will appear here.$>$}

\def\authorrefmark#1{\ensuremath{^{\textbf{#1}}}}

\makeatletter
\renewcommand{\p@subsection}{\thesection-}
\makeatother

\begin{document}
\receiveddate{XX Month, XXXX}
\reviseddate{XX Month, XXXX}
\accepteddate{XX Month, XXXX}
\publisheddate{XX Month, XXXX}
\currentdate{XX Month, XXXX}
\doiinfo{XXXX.2022.1234567}

\markboth{}{Wang and Yang}

\title{Rethinking Multi-Branch and Cross-Backbone Fusion for Vehicle Re-Identification under Foundation-Model Pretraining}

\author{Yu Wang\authorrefmark{1} and Hongyu Yang\authorrefmark{1}}
\affil{Huahuan (Yunnan) Technology Co., Ltd., Yunnan, China}
\corresp{Corresponding authors: Yu Wang (wangyu@cn-es.com) and Hongyu Yang
(yanghongyu@cn-es.com). \textbf{Author contributions:} Y.W.: conceptualization,
methodology, software, experiments, analysis, original draft; H.Y.: supervision,
validation, review and editing. Both authors approved this submission.
\textbf{Funding:} none. \textbf{Competing interests:} none declared in any method
evaluated. \textbf{Generative AI:} used for language editing and for scripting
parts of the analysis pipeline; all experimental design, results and conclusions
are the authors' own, and every reported number was produced by the released code.
\textbf{Code and data availability:} code, the branch-level diagnostic toolkit,
the exact sparse re-ranking implementation and the Option~A checkpoint are
released at \url{https://gitee.com/robertwangwang/Dinov3ForMBR}. Both benchmarks
\cite{veriwild,veri776} are public but distributed under research-use agreements
precluding redistribution, so the released code reproduces every analysis in
Section~\ref{sec:stress} once the data are obtained.}

\begin{abstract}

    Multi-branch architectures and CNN--Transformer fusion are widely believed to
improve vehicle re-identification (Re-ID) by combining complementary
representations. We revisit this for a DINOv3-pretrained backbone. A \emph{single}
DINOv3-pretrained ConvNeXt with a tuned recipe reaches \textbf{88.19 mAP on
VeRi-Wild Small and 77.47 on Large from visual cues alone}, within the combined
evaluation and optimization noise of the strongest protocol-verified
metadata-dependent multi-branch baseline, and \textbf{92.38/83.68 with
training-free re-ranking}. Using this baseline and retrieval-level branch
diagnostics, we ask whether representational diversity still pays at this scale.
In our runs, it does not. Across both benchmarks and every converged configuration,
concatenating multiple heads over a shared backbone moves the best single head by
under one mAP point in either direction while costing four times the embedding
dimension; 99.7\% of the concatenation's variance lies in 512 principal
components, so the heads not only duplicate one another but each occupies a
quarter of its nominal 2048 dimensions. Pushing diversity to its architectural
limit, CNN versus Transformer, we grant fusion every advantage through an
\textbf{asymmetric frozen-anchor scheme}. Every Transformer configuration still
lands at least 13 mAP below the ConvNeXt backbone (13--15 for the two strongest,
up to 46 for the weakest), and a paired per-query bootstrap
bounds the fusion gain at \textbf{$+0.11$ mAP (95\% CI)} even for the most
favourable snapshot we obtained. One strong backbone with the right recipe and
re-ranking is the efficiency frontier. All results use \emph{single-seed} training
and one foundation-model family; differences of this size are therefore reported
as bounds rather than orderings, and we list falsifiers.

\end{abstract}

\begin{IEEEkeywords}
Cross-camera evaluation, feature fusion, foundation models, multi-branch
networks, re-ranking, representation similarity, vehicle re-identification,
vision transformers.
\end{IEEEkeywords}


\maketitle

\section{Introduction}
\IEEEPARstart{V}{ehicle} re-identification (Re-ID)---matching a query vehicle
across non-overlapping cameras---underpins cross-camera tracking, traffic
analytics and urban surveillance \cite{survey,veriwild}. For nearly a decade its
progress has been driven by \emph{representational diversity}: multi-branch and
multi-granularity networks split features into parallel streams \cite{mgn,mbr},
part-based modules extract local descriptors \cite{transreid}, and a growing line
fuses heterogeneous backbones---typically a CNN with a Transformer---on the
premise that complementary inductive biases yield cues neither captures alone (a
premise vehicle Re-ID imports from person Re-ID \cite{fusionreid}). The shared,
rarely questioned assumption is that \emph{more diverse representations are
better}: each branch contributes unique correct matches, and their aggregation
surpasses any single stream \cite{mbr,mgn}.

That assumption was established with ImageNet-pretrained, lightweight backbones,
where no extractor was near saturation and diversity was a cheap path to
coverage. Foundation models change the premise: DINOv3 rivals specialized state
of the art \emph{without} fine-tuning \cite{dinov3}, and in Re-ID merely
fine-tuning a foundation encoder is already competitive \cite{clipreid}, leaving
little headroom for extra branches---while fine-tuning is also known to
\emph{distort} pretrained features \cite{lpft}, casting doubt on whether branch
diversity survives joint optimization at all. That branches \emph{can} be
redundant is itself established, motivating complementarity constraints and
redundancy-reduction modules \cite{cran,hbmcn,kminet}; but those results predate
foundation-model backbones and argue from loss- and feature-level surrogates or
end-to-end ablations. Whether the premise survives at this scale, measured where
it must pay off (retrieval), has to our knowledge not been tested---yet it
governs how practitioners spend compute: on one strong backbone, or several fused
ones.

We re-examine it on VeRi-Wild \cite{veriwild} and VeRi-776 \cite{veri776} under
the official cross-camera protocol, whose omission inflates reported numbers by
3--4 mAP. We first establish that a \emph{single} DINOv3-ConvNeXt
\cite{dinov3,convnext}, given a tuned recipe, reaches parity with the strongest
protocol-verified metadata-dependent multi-branch baseline---echoing, at
this pretraining scale, that training procedure can matter as much as architecture
\cite{rsb,bot}. Against it we stress-test diversity with a reusable diagnostic
toolkit, and do not stop at a na\"ive fusion that could be dismissed as
under-engineered: we diagnose why symmetric joint fine-tuning of two foundation
backbones collapses---a learning rate tolerable for one erodes both, and the
apparent $+4.9$ gain is measured against a jointly \emph{degraded} base---then
give fusion every advantage through an \textbf{asymmetric frozen-anchor scheme}
that freezes the 88.19 ConvNeXt and trains only the Transformer and fusion module,
under \textbf{three recipes plus a LoRA control}.

Our findings are consistently negative for the premise---and that, we argue, is
the contribution. Heads over a shared backbone become \emph{more} alike with
training, consistent with solutions on one optimization trajectory occupying a
single function-space mode, whereas ensemble diversity classically derives from
independent training \cite{ensembles-landscape,deep-ensembles}. Cross-backbone
heterogeneity is genuine and \emph{survives} training (top-10 Jaccard 0.54 against
0.79--0.84 for same-backbone heads), in line with documented ViT/CNN differences
\cite{vit-vs-cnn}---yet it resisted aggregation: the Transformer did not cross a
${\sim}73$ mAP ceiling under any full-fine-tuning recipe we tried, and an oracle fusion of our
\emph{strongest} Transformer assigns it weight zero. Diversity is not sufficient;
it must also be aggregable, and here it was not.

\noindent\textbf{A signal-processing reading.} Though posed on images, the objects
here are signal-processing ones, and the negative result concerns
\emph{aggregation operators} rather than architectures. A multi-branch embedding is
a uniformly weighted \emph{linear combination of similarity kernels}:
concatenating unit-normalized descriptors and renormalizing computes the
arithmetic mean of the per-branch cosine kernels (Section~\ref{sec:prelim}), so
asking whether branches help is asking when such a combination beats its best
constituent kernel---and our oracle analysis answers the corresponding
\emph{optimal-weighting} question. The PCA probe is a \emph{spectral} statement
about that aggregate; the re-ranking result a memory-bounded execution of an
\emph{affinity-matrix} computation, replacing a dense $N\times N$ operator with an
inverted-index form that provably preserves its output; and the diagnostics are
\emph{agreement measures between ranking functions}, which is what a fusion rule
operates on. Re-ID has an established presence in this community---in
retrieval/verification loss design \cite{jstsp-reid} and graph-based video
matching \cite{spl-reid}---and the finding is portable wherever learned kernels
are averaged in the hope their disagreements accumulate.

We contribute:

\begin{itemize}
\item \textbf{(i)} the first \emph{retrieval-level} evidence that the diversity
premise fails under fine-tuning at this pretraining scale, in two ways. Same-backbone
heads collapse into redundancy, replicated in-domain on \emph{both} benchmarks:
across six VeRi-776 configurations and every converged VeRi-Wild configuration,
concatenation moves the best single head by under one mAP point, though the
\emph{sign} depends on feature resolution (Table~\ref{tab:veri776})---so we state
a bound, not a point estimate. Redundancy has precedent \cite{cran,hbmcn,kminet};
what is new is that it is measured at retrieval, that it is the \emph{endpoint} of
shared optimization rather than a remediable defect, and that cross-backbone
diversity \emph{survives} training yet resisted \emph{aggregation}---an oracle
score-level fusion of our strongest ViT assigns it weight zero (established under
VeRi-Wild training);
\item \textbf{(ii)} a documented recipe lifting a single DINOv3-ConvNeXt to parity
with the strongest protocol-verified metadata-dependent multi-branch baseline
(MBR-4B-LAI) on VeRi-Wild;
\item \textbf{(iii)} a branch-level diagnostic toolkit---complementing
representation-level similarity \cite{cka,wide-deep} with retrieval-level
measures---and a convergence-only measurement discipline that matters,
as undertrained snapshots can systematically mislead;
\item \textbf{(iv)} an efficiency frontier built on a
\emph{memory-bounded, output-identical} execution of $k$-reciprocal re-ranking
\cite{kreciprocal}. Memory-scalable re-ranking is not new---prior work reduces
$O(n^2)$ to $O(kn)$ \cite{bertocco} or restructures it via feature propagation
and adaptive neighbour search \cite{gcr,chebgr}---but each \emph{changes the
output}. Ours reproduces the classical algorithm bit-for-bit ($\Delta$mAP
$=0.0000$), so published re-ranking numbers transfer unchanged, while clearing
the memory wall that makes the classical form unrunnable at 128k gallery scale.
\end{itemize}

\section{Related Work}

\noindent\textbf{Vehicle Re-ID.} The field has matured from metric learning on
VehicleID \cite{vehicleid} to in-the-wild benchmarks such as VeRi-776
\cite{veri776} and VeRi-Wild \cite{veriwild}, with recent methods exploiting
multi-branch topologies, viewpoint/attribute metadata and Transformer backbones
\cite{survey}. The line remains active: 2025--26 designs still pair global with
local branches \cite{kminet,lrgsnet} or add a metadata-conditioned third branch
\cite{sednet}. Strong inter-class similarity between same-model vehicles and
severe viewpoint variation have made \emph{representational diversity} an
attractive design principle here: precisely the premise we revisit.

\noindent\textbf{Multi-branch and multi-granularity Re-ID.} Splitting features
into parallel streams is dominant: MGN \cite{mgn} combines global with partitioned
local branches; MBR \cite{mbr} pairs a ResNet50-IBN global branch with a BoT
self-attention branch under Loss-Branch-Split, attributing its gains to ``improved
feature diversity''; TransReID \cite{transreid} adds jigsaw patch rearrangement;
FusionReID \cite{fusionreid} (a \emph{person} method whose premise the vehicle
literature inherits) fuses CNN and Transformer features through stacked
mutual-attention. Bag-of-Tricks \cite{bot} is an early counterpoint: a single
global stream with careful training rivals multi-branch designs.

\noindent\emph{Branch redundancy is not a new observation, and we do not claim it
as one.} It is already recognised as a defect to engineer around: CRAN
\cite{cran} opens from ``severe redundancy'' among parallel attention branches and
imposes complementarity and orthogonality constraints; HBMCN \cite{hbmcn} compares
heterogeneous against homogeneous branches (both person Re-ID); and in vehicle
Re-ID, KMINet \cite{kminet} adds an attention-transfer module specifically to
``reduce feature redundancy'' across branches. What this line shares is an
assumption we test rather than adopt---that redundancy is \emph{remediable}. Three
things remain open, and are what we contribute: \emph{(a)} that literature
validates diversity through end-to-end mAP ablations or loss- and feature-level
surrogates, whereas we measure it where it must pay off, at \emph{retrieval};
\emph{(b)} those results predate foundation-model backbones, and we find that with
a near-saturated trunk redundancy is not a defect to constrain away but the
\emph{endpoint} of a shared optimization trajectory; \emph{(c)} where diversity
does survive---across backbones---it is nonetheless \emph{not aggregable}, which
no redundancy-reduction method addresses.

\noindent\textbf{Foundation models and fine-tuning.} DINOv3 \cite{dinov3} shows
a single self-supervised backbone can match specialized state of the art without
fine-tuning, and CLIP-ReID \cite{clipreid} shows that merely fine-tuning a
foundation encoder is already competitive in Re-ID; CLIP-based semantic
enhancement reports among the highest published VeRi-Wild figures
\cite{clipsenet}, though not under a protocol we could verify
(Section~\ref{sec:parity}). Kumar \emph{et al.} \cite{lpft} prove that full
fine-tuning distorts pretrained features, motivating freezing strategies, with
layer-wise LR decay---as applied to self-supervised pretraining in
\cite{beit}---the standard mitigation. What joint fine-tuning does to
\emph{multi-branch} topologies built atop such backbones has not been examined;
in our setting it homogenizes them.

\noindent\textbf{Representation similarity and ensemble diversity.} CKA
\cite{cka} and its applications \cite{wide-deep,vit-vs-cnn} characterize when
networks learn similar representations---notably that ViTs and CNNs differ
substantially, and that equally accurate models can still err differently. Deep
ensembles \cite{deep-ensembles} derive their gains from independently trained
members, and Fort \emph{et al.} \cite{ensembles-landscape} show that solutions
along a shared optimization trajectory cluster in one function-space mode while
independent initializations explore distinct ones. Our diagnostics extend this
literature to \emph{retrieval-level} measures, and our findings instantiate the
shared-trajectory prediction in the multi-branch Re-ID setting.

\section{Evaluation Protocol and Preliminaries}
\label{sec:prelim}

\noindent\textbf{Datasets and splits.} VeRi-Wild \cite{veriwild}: we report the
official test\_3000 split (3,000 query IDs, 38,861 gallery images; ``Small'')
and test\_10000 (128,517 gallery; ``Large''). VeRi-776 \cite{veri776} serves as
a \emph{cross-domain} target: models trained on VeRi-Wild are evaluated on
VeRi-776 without seeing it.

\noindent\textbf{Protocol discipline.}\label{sec:protocol} All numbers use the
official cross-camera evaluation: gallery images sharing both the query's vehicle
ID and camera are excluded as junk, along with the query itself. Measured on our
own dual-backbone reproductions, omitting that filter inflates mAP from 82.42 to
85.46 on VeRi-776 (\textbf{$+3.04$}) and from 73.89 to 77.83 on VeRi-Wild
(\textbf{$+3.94$}). We verified our implementation reproduces the original MBR
evaluation code exactly, and all comparisons here are protocol-matched. We flag
this as a reproduction pitfall---inflated, unfiltered numbers land close enough
to filtered published figures to cause silent misranking---rather than as a claim
that any specific published pipeline omitted the filter.

\noindent\textbf{Concatenation arithmetic, and where the headroom comes from.} For
unit-normalized branch embeddings $f_1,\dots,f_B$, stacking and renormalizing by
$\sqrt{B}$ gives $F=\tfrac{1}{\sqrt{B}}[f_1;\dots;f_B]$, so
$\langle F_a,F_b\rangle=\tfrac{1}{B}\sum_i\langle f_{i,a},f_{i,b}\rangle$---the
\emph{arithmetic mean} of the per-branch cosines. It follows that if every branch
induces the same gallery ranking, that mean is an order-preserving combination of
identical orderings and reproduces the ranking exactly; since mAP and CMC depend
only on the ranking, concatenation then equals every branch and cannot exceed the
best. A strict gain requires branches to \emph{disagree} on at least one query.
Elementary, but it fixes what must be measured: inter-branch disagreement at
retrieval \emph{is} the headroom available to concatenation---what the
diagnostics of Section~\ref{sec:stress} target and the oracle bound of
Section~\ref{sec:crossbackbone} exhausts.

\section{A Single Foundation Backbone Is (Almost) All You Need}
\label{sec:main}

\subsection{Architecture and recipe}

Our model (``Option~A'') is deliberately \emph{not} novel: a single
DINOv3-distilled ConvNeXt-Base backbone \cite{dinov3,convnext} under MBR's
Loss-Branch-Split topology \cite{mbr}---four heads (conv-cls, conv-metric, and
two Bottleneck-Transformer heads BoT-cls, BoT-metric \cite{bot}), each producing
a 2048-d embedding concatenated to 8192-d---trained with cross-entropy and metric
losses \cite{circle,center,triplet}. What is tuned is the \emph{recipe}:
full-depth features, full unfreezing after a frozen warm-up, full training data,
and staged LR decay ($\times 0.1$ at epoch 40). Training runs on 8 NVIDIA RTX PRO
6000 GPUs with P$16$K$4$ sampling (effective batch $16\times4\times8=512$).

Each recipe component maps to a step-like jump: frozen Phase 1 ends at 39.46 mAP;
unfreezing jumps to 56.25 in one epoch ($+16.8$); a steady climb to 85.98 by
epoch 31; the epoch-40 LR decay adds \textbf{$+2.21$ (VeRi-Wild)}, converging at
\textbf{88.19} (epoch 42). A component ablation (Section~\ref{sec:samebackbone})
attributes most of this gain to the LR-decay stage and shows the full-depth
choice is \emph{dataset-dependent} (it reverses on VeRi-776).

\noindent\textbf{Training cost, and how the recipe was chosen.} For the run we
instrumented end to end---the dual-backbone system---an epoch takes ${\sim}25$ min
frozen and ${\sim}35$ min in the main stage (${\sim}35$ h for 60 epochs);
Option~A is cheaper by roughly the FLOP ratio of Section~\ref{sec:efficiency}. We
did not instrument the recipe search's cumulative budget and report that as a gap.
More consequential: \textbf{the recipe was selected against test-split mAP}, since
neither benchmark defines a held-out validation split and the baselines we compare
against are tuned the same way. Absolute recipe-level figures therefore
\emph{characterize} these systems rather than being held-out estimates---but this
does not propagate into the paper's central comparisons, which are made
\emph{within} a fixed recipe and run (branch against branch, fused against its own
single backbone) and are
unaffected by where that recipe sits in absolute terms.


\subsection{Results: metadata-free matches metadata-dependent}
\label{sec:parity}

Table~\ref{tab:veriwild} reports our main VeRi-Wild results, protocol-matched to
\cite{mbr}. Our margins over the strongest metadata-dependent variant MBR-4B-LAI
are under a tenth of a point on both splits---far smaller than the evaluation's
own uncertainty: a query-level bootstrap (2,000 resamples) gives \textbf{$\pm 0.66$
mAP on Small (88.19) and $\pm 0.51$ on Large (77.47)}, wider than that margin by
nearly an order of magnitude. We therefore claim \emph{parity, not superiority},
scoped to the protocol-verified comparison: purely visual features match the
strongest metadata-dependent configuration of the strongest published
multi-branch baseline, removing camera/view annotations from the requirements
list. CMC-1 reaches 96.39 / 91.86. The interval covers \emph{evaluation}
uncertainty only; these are single-seed runs, so parity is a statement about our
runs and we claim no ordering in either direction.

\begin{table}[!t]
\caption{VeRi-Wild results, split by whether we could verify the evaluation
protocol. Purely visual Option~A reaches parity with the strongest
\emph{protocol-verified} metadata-dependent MBR variant. The lower block lists
recent (2025--26) methods---two multi-branch designs \cite{kminet,lrgsnet} and a
three-branch design \cite{sednet}---to show the comparison set is current;
``n.v.'' marks figures we did not tabulate, being unable to confirm their split or
filtering. SEDNet's third branch consumes direction and viewpoint metadata, i.e.\
the per-image annotation dependency our system removes.}
\label{tab:veriwild}
\centering
\begin{tabular}{lcccc}
\hline
Method & Metadata & Small mAP & Large mAP & Protocol \\
\hline
\multicolumn{5}{l}{\emph{Protocol-verified (reproduced against MBR evaluation code)}}\\
MBR-4B \cite{mbr} & none & 87.97 & 77.15 & verified \\
MBR-4B-LAI \cite{mbr} & cam.+view & 88.12 & 77.41 & verified \\
\textbf{Option A (ours)} & \textbf{none} & \textbf{88.19} & \textbf{77.47} & verified \\
\textbf{~+ re-ranking} & none & \textbf{92.38} & \textbf{83.68} & verified \\
\hline
\multicolumn{5}{l}{\emph{Recent methods, protocol not verifiable}$^{\dagger}$}\\
CLIP-SENet \cite{clipsenet} & none & 89.1 & --- & self-rep. \\
KMINet \cite{kminet} (multi-branch) & none & 88.56 & --- & self-rep. \\
LRGS-Net \cite{lrgsnet} (multi-branch) & none & n.v. & n.v. & self-rep. \\
SEDNet \cite{sednet} (3-branch) & dir.+view & n.v. & n.v. & self-rep. \\
\hline
\end{tabular}

\vspace{2pt}
\begin{flushleft}
\footnotesize $^{\dagger}$As reported by the respective authors; splits are not
always identified, implementations are generally unavailable, and we could not
verify cross-camera junk filtering---which inflates mAP by
\textbf{+3.04}/\textbf{+3.94} on our own system (Section~\ref{sec:protocol}),
several times the ${\le}0.9$-point spread across this block. These figures are
therefore not established to be comparable in either direction; we claim parity
only with the strongest \emph{protocol-verified} baseline. KMINet's self-reported
88.56 falls \emph{inside} our own evaluation interval for 88.19 ($\pm0.66$): even
at face value it would not establish an ordering---the reading we ask for our own
numbers.
\end{flushleft}
\end{table}

\noindent\textbf{Cross-domain generalization.} Zero-shot on VeRi-776, Option~A
reaches \textbf{66.02 mAP / 87.07 R1} against the matched in-domain reference of
84.07 (Section~\ref{sec:samebackbone}); the $-18.0$ gap with an \emph{intact} R1
is consistent with fine-tuning's known out-of-distribution cost \cite{lpft}.

\subsection{Exact sparse re-ranking at scale}

$k$-reciprocal re-ranking \cite{kreciprocal} is the deployment workhorse
($+4.19$ Small / \textbf{$+6.21$} Large, R1 preserved), but its classical
implementation is memory-bound: on test\_10000 it needs ${\sim}190$ GB and fails
outright. Scaling re-ranking is an active line: Bertocco \emph{et al.} \cite{bertocco} cut memory from $O(n^2)$ to $O(kn)$ and
evaluate on VeRi-Wild; GCR \cite{gcr} recasts re-ranking as feature propagation;
Cheb-GR \cite{chebgr} replaces $k$-NN search with an adaptive neighbourhood rule.
\textbf{Ours is orthogonal: those change the output, ours is bit-identical---and
it is an implementation result, not an algebraic one.} With row
sums $s_i=\sum_p v_{ip}$ and overlaps $M_{ij}=\sum_p\min(v_{ip},v_{jp})$,
$\max(a,b)=a+b-\min(a,b)$ gives $d_J(i,j)=1-M_{ij}/(s_i+s_j-M_{ij})$. We claim no
novelty for this: the inverted-index accumulation of $\sum_p\min$ already appears
in the implementation released with \cite{kreciprocal}, which normalizes $v$
row-wise and therefore specializes to $1-M/(2-M)$. What that implementation still
materializes is the dense $N\times N$ distance and Jaccard matrices, and that---not
the Jaccard sum---is the source of the ${\sim}190$ GB. Our contribution is the
\emph{execution}: general row sums, a blocked top-$k_1$ selection replacing the
full $N\times N$ argsort, and a query-block-only $M$, giving $O(Nk_1+n_qN)$ peak
memory with \emph{no term dropped}. Empirically: identical retrieval on
test\_3000 ($\Delta$mAP $=0.0000$), and 29 s vs.\ 534 s---an
\textbf{$18\times$} we attribute to the implementation as a whole, since the
blocked selection runs on GPU where the classical argsort runs on CPU; exactness
and memory do not depend on it. test\_10000 completes in 137 s within tens of GB (one RTX 3090, standard
$k_1=20$, $k_2=6$, $\lambda=0.3$ \cite{kreciprocal}). Larger galleries benefit
\emph{more} from re-ranking, so removing the memory wall matters precisely where
the gain is largest.

\section{Stress-Testing the Diversity Premise}
\label{sec:stress}

\subsection{Branch-level diagnostics}
\label{sec:diagnostics}

Guided by the remark of Section~\ref{sec:prelim}, we measure disagreement where it
counts---retrieval behavior---with four instruments: \textbf{per-branch
retrieval} (each branch's standalone mAP/R1: how strong is each stream?);
\textbf{top-$k$ Jaccard} (overlap of two branches' top-10 retrieval sets: do they
return the same gallery images?); \textbf{occlusion-saliency agreement} (Pearson
correlation and top-20\% IoU of occlusion-sensitivity heatmaps \cite{zeiler}: do
they \emph{look at} the same regions?); and \textbf{complementary-correctness}
(the fraction of queries one branch ranks correctly and the other does not: is
there unique signal to harvest?). These complement representation-level tools such
as CKA \cite{cka}: two branches can be CKA-dissimilar yet retrieve identically, so
only retrieval-level measures bound fusion headroom directly.

\subsection{Same-backbone multi-head: collapse into redundancy}
\label{sec:samebackbone}

At convergence (E42) the four heads read 88.11, 88.13, 88.24 and 88.07 mAP---a
spread of 0.17---and concatenation lands at 88.19, delivering \textbf{no
statistically distinguishable gain} over the best single head: a \emph{paired}
per-query bootstrap of the difference gives $\Delta$(concat $-$ BoT-cls)
$= \textbf{$-0.05$ mAP}$, 95\% CI $[-0.19,+0.09]$, covering zero. The paired form
matters here---the corresponding \emph{unpaired} single-model CI has half-width
$\pm 0.66$, far too wide to adjudicate a 0.05 difference; pairing cancels the
shared query-set variance and shrinks the interval nearly fivefold. The diversity
trajectory (Table~\ref{tab:diversity}) explains why the heads are redundant.

\noindent\textbf{In-domain replication on VeRi-776.} The collapse is not specific
to VeRi-Wild. We train the \emph{same} Option~A in-domain on VeRi-776 (575 IDs) in
six configurations---one reference system matched to the VeRi-Wild recipe, plus a
five-cell leave-one-out ablation at this benchmark's classical batch size---and
report \emph{all} of them in Table~\ref{tab:veri776}. The reference row carries
the replication claim: concatenation (84.07) again fails to beat the best single
head (84.69), $\Delta=-0.62$, reproducing VeRi-Wild's $-0.05$ and slightly
exceeding it.

\noindent\emph{$\Delta$ is configuration-dependent, and we report that rather than
average over it.} Across the six runs $\Delta$ spans $-0.76$ to $+0.58$: four are at or below zero
and two are positive---A1 ($+0.58$, paired 95\% CI $[+0.49,+0.68]$) and A4
($+0.24$). Both positives sit where a head is \emph{furthest from saturation}: A1
doubles the resolution fed to the heads ($8{\times}8\rightarrow16{\times}16$), and
A4 is the weakest cell in the table (best head 75.71). Where heads are strong and see
identical coarse evidence the residue vanishes; where they are resolution- or
capacity-limited some genuine complementarity survives---the main thesis, not a
contradiction to it. Even at its most favourable that residue
is an order of magnitude below the $+2$ to $+4$ the literature reports. Hence a \emph{bound} rather than a point estimate:
the sign is not stable across configurations, but the magnitude is.

\noindent\emph{Recipe attribution.} Reading down the concat column, the dominant positive levers are the \emph{ConvNeXt}
heads (substituting ResNet-Bottleneck heads \textbf{costs 6.37 mAP},
$82.32\rightarrow75.95$) and staged LR decay (removing it costs 5.88); the
Loss-Branch-Split assignment is neutral ($-0.01$), against our own prior
expectation. One component reverses outright:
\textbf{full-depth features are harmful on VeRi-776}, where truncating the trunk
one stage earlier---doubling the spatial resolution reaching the
heads---\emph{gains} 5.00 mAP ($82.32\rightarrow87.32$). On VeRi-Wild the same
substitution is neutral (88.69 vs.\ 88.19, runs differing in batch and schedule,
so comparable rather than ordered). Semantic depth is therefore not the lever the
recipe search first suggested; we retain the full-depth system as reference only
because every analysis below is anchored on it.

\begin{table}[!t]
\caption{\textbf{Every VeRi-776 Option~A run in this paper, on one axis}; all
share one architecture.
\textbf{Row 1 is the VeRi-776 reference system}---matched to the VeRi-Wild recipe
and batch---and supports the replication claim. A0--A4 are a leave-one-out
ablation at this benchmark's classical effective batch of 128, so their
\emph{absolute} values are not comparable to row 1; only \emph{differences within}
the block are. $\Delta =$ concat $-$ best head (from unrounded values) spans $-0.76$ to
$+0.58$---under one mAP point either way, but with a configuration-dependent
\emph{sign}, hence the bound in contribution~(i). All values come from the \emph{final
post-decay} checkpoint, never the best epoch, which here sits on a pre-decay spike
and inflates recipe deltas by up to $24\times$. The 82.42 of Section~\ref{sec:protocol} is excluded by
design (a dual-backbone model kept only to quantify protocol inflation); its
closeness to A0's 82.32 is coincidental.}
\label{tab:veri776}
\centering
\resizebox{\columnwidth}{!}{%
\begin{tabular}{llccccc}
\hline
Configuration & Trunk & Batch & Sched. & Best head & Concat & $\Delta$ \\
\hline
\textbf{Reference (VeRi-Wild-matched)} & full & 512 & 60 ep & 84.69 & \textbf{84.07} & $-0.62$ \\
\hline
A0 ablation base & full & 128 & 50 ep & 82.35 & 82.32 & $-0.03$ \\
A1 $-$ full depth (truncated) & trunc. & 128 & 50 ep & 86.74 & \textbf{87.32} & $\mathbf{+0.58}$ \\
A2 $-$ staged LR decay & full & 128 & 50 ep & 76.66 & 76.44 & $-0.22$ \\
A3 $-$ Loss-Branch-Split & full & 128 & 50 ep & 83.10 & 82.33 & $-0.76$ \\
A4 ResNet heads (not ConvNeXt) & full & 128 & 50 ep & 75.71 & 75.95 & $+0.24$ \\
\hline
\end{tabular}}
\end{table}

\begin{table}[!t]
\caption{Diversity metrics against training, with bootstrap 95\% intervals
(2,000-fold). Same-backbone heads begin apart and \emph{converge}; cross-backbone
pairs stay genuinely heterogeneous, and the two groups' intervals do not overlap.
Occlusion agreement is measured only at convergence.}
\label{tab:diversity}
\centering
\resizebox{\columnwidth}{!}{%
\begin{tabular}{lccc}
\hline
Metric & Early & Converged (E42) & Sample \\
\hline
\multicolumn{4}{l}{\emph{Same backbone} (conv $\leftrightarrow$ BoT heads)}\\
top-10 Jaccard (cls/metric) & 0.64 (E1) $\rightarrow$ 0.79 (E3) & \textbf{0.835/0.841} $\pm0.005$ & 3000 queries \\
Occlusion Pearson & --- & \textbf{0.96} [0.95, 0.97] & 200 image pairs \\
Occlusion top-20\% IoU & --- & \textbf{0.81} [0.79, 0.83] & 200 image pairs \\
\multicolumn{4}{l}{\emph{Cross backbone} (ConvNeXt $\leftrightarrow$ ViT-L, frozen)}\\
Occlusion Pearson & --- & \textbf{0.30} [0.26, 0.34] & 200 image pairs \\
Occlusion top-20\% IoU & --- & \textbf{0.20} [0.18, 0.22] & 200 image pairs \\
\hline
\end{tabular}}
\end{table}

Heads grow \emph{more} alike as training proceeds---the opposite of the
diversity premise, and consistent with shared-trajectory solutions collapsing
into a single function-space mode \cite{ensembles-landscape}. A PCA probe
delivers the structural verdict: \textbf{99.7\% of the 8192-d concatenation's
variance lies in 512 principal components}, with mAP unchanged at 88.20
(projections to 1024 and 2048 are likewise lossless). We are deliberate about
what this does and does not say. Four \emph{identical} 2048-d heads would already
span only 2048 dimensions, so the drop from 8192 to 2048 is exactly what branch
duplication alone predicts; the \emph{further} collapse to 512 is a property of a
single head, which therefore occupies only a quarter of the dimensions it
nominally carries. The concatenation is redundant twice over---across heads, and
within each head---and only the first of those is attributable to the
multi-branch design.

\subsection{Cross-backbone fusion: maximum heterogeneity, zero net gain}
\label{sec:crossbackbone}

\noindent\textbf{Genuine heterogeneity at the starting line.} With both backbones
frozen, ConvNeXt and ViT-L occlusion saliencies correlate at only \textbf{0.30}
(95\% CI [0.26, 0.34])---the lowest we measure anywhere against 0.96 for
same-backbone heads, with non-overlapping intervals
(Table~\ref{tab:diversity})---confirming that CNN/Transformer differences
\cite{vit-vs-cnn} survive DINOv3 distillation at initialization.
Complementary-correctness reaches 28--32\%. If diversity pays anywhere, here.

\noindent\textbf{Symmetric joint fine-tuning fails---and how it fails matters.}
Our dual-backbone model (ConvNeXt full-depth + ViT-L \cite{vit}, two
cross-attention fusion modules, learned fusion head, ${\sim}560$M training-time
params, ${\sim}404$M deployed) peaks at 85.30 (E7) then \emph{declines for ten
straight epochs} to 83.36. The reversal coincides exactly with the LR schedule
reaching its 2.5e-4 plateau and with synchronous loss spikes, pinning the collapse
to optimization rather than architecture; restarting at LR$\times 0.1$ immediately
recovers to \textbf{86.32}. At that rescued state we observe the \emph{only
genuine fusion gain in this study}: fused 86.32 vs.\ best
single branch (cnx\_g) 81.38, \textbf{$+4.94$}, Jaccard 0.535. But the comparison
base is contaminated---joint training had degraded that ConvNeXt to 81.38, far
below the 88.19 the \emph{same} backbone reaches alone---and the fused
86.2--86.3 never approaches 88.19.

\begin{figure}[!t]
\centering
\includegraphics[width=\columnwidth]{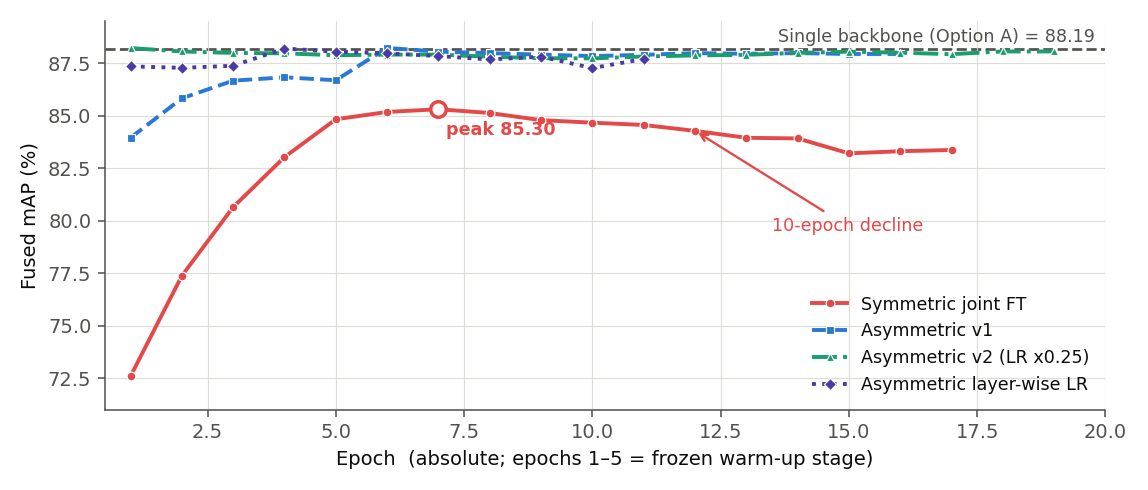}
\caption{Fused mAP vs.\ epoch. The symmetric run climbs to 85.30 then declines for
ten straight epochs; every asymmetric frozen-anchor run holds at parity with the
single backbone (dashed, 88.19). The asymmetric peaks (up to 88.22, v1 at absolute
epoch 6) occur at the start of the main stage---already matching the 8192-d
single-backbone system with a 1024-d fused output---and settle to 87.7--88.05 as
the ViT integrates: anti-collapse and parity, never a meaningful gain. v3
(3 epochs) is omitted here but tabulated in Table~\ref{tab:recipes}.}
\label{fig:anticollapse}
\end{figure}

\noindent\textbf{Asymmetric frozen-anchor: giving fusion every advantage.} We
therefore transplant the converged 88.19 ConvNeXt \emph{system} (all 350 weight
keys) and \emph{freeze} it as a semantic anchor, training only the ViT, fusion
modules and fusion head. Two quantities must be kept distinct: the anchor is the
full 8192-d Option~A embedding (88.19), whereas \emph{cnx\_g} is a diagnostic
1024-d pooled probe of the frozen trunk reaching only 86.5---an untrained pool,
not a head---so we do \emph{not} claim a by-construction floor of 88.19. What the
anchor \emph{does} guarantee is anti-collapse: on the same 2.5e-4 plateau that
coincided with the symmetric run's slide, all asymmetric runs hold stable across
ViT rates 2.5e-5--1.125e-4 (Fig.~\ref{fig:anticollapse}), isolating the anchor
rather than any particular ViT rate as the stabilizer. To preempt the objection
that the ViT was under-trained we train it under three full-fine-tuning recipes
and, because full fine-tuning of self-supervised ViTs is a documented instability
that parameter-efficient adaptation mitigates \cite{lora,peft-ssl-vit,lpft}, add a
LoRA control.

\begin{table*}[!t]
\caption{ViT-branch mAP for every asymmetric frozen-anchor run, plus a standalone
ViT trained \emph{outside} the fusion topology and a LoRA (PEFT) control. Full
fine-tuning caps at ${\sim}73$ while LoRA reaches 75.19, so ${\sim}73$ is partly
fragility, not a hard wall; the robust result is that \emph{every} configuration
stays ${\ge}13$ below the ConvNeXt backbone (88.19). Epochs count \emph{within the
main stage}, which follows a five-epoch frozen warm-up (absolute epoch $=$ label
$+5$). \textbf{v1 carries the global maximum of fused mAP over all runs and epochs
(88.22, main-stage E1 $=$ absolute epoch 6)---the snapshot the $+0.11$ paired
bound is computed on}; its ViT reads 42.3 in that same snapshot, and v2,
warm-started from v1, inherits that state as its origin, so the two rows share one
measurement rather than reporting two. ``anchor probe'' is a 1024-d pooled probe
of the frozen trunk, \emph{not} the transplanted 8192-d Option~A system; as an
untrained pool it reads 86.5, below a trained head. The LoRA control is not
\emph{trained} under the anchor, but oracle score-level fusion assigns it weight
zero without training.}
\label{tab:recipes}
\centering
\resizebox{\textwidth}{!}{%
\begin{tabular}{lcccc}
\hline
Recipe / control & vit\_g trajectory & vit\_g peak & fused (peak $\rightarrow$ final) & anchor probe (cnx\_g) \\
\hline
\textbf{v1: uniform LR $\times 0.1$ (2.5e-5)} & 13.1 (warm-up) $\rightarrow$ 42.3 (E1) & 42.3 (at fused peak) & \textbf{88.22} $\rightarrow$ 87.93 & 86.47 \\
v2: uniform LR $\times 0.25$ (6.25e-5), \emph{from v1} & (v1 state) $\rightarrow$ 70.7 (E4) $\rightarrow$ 73.1 (E18) & \textbf{73.1 (plateau)} & 88.21 $\rightarrow$ 88.05 & 86.46 \\
v3: uniform LR $\times 0.45$ (1.125e-4) & 64.2 (E2), stopped E3 & \textbf{64.2} (E3) & 88.21 $\rightarrow$ 87.98 & 86.50 \\
Layer-wise LR decay (as in \cite{beit}; peak 3e-4, 0.98/layer) & 67.4 (E2) $\rightarrow$ \textbf{68.9 (E3 peak)} $\rightarrow$ 59.7 (E8) & ${\sim}69$, then \textbf{erodes} & 88.19 $\rightarrow$ 87.69 & 86.47 \\
\emph{Control:} standalone ViT (no fusion, recipe v2) & plateau at E26 & \textbf{67.0} (72.1 no-cam) & n/a (no fusion) & --- \\
\emph{Control:} LoRA ViT branch (rank 8, $\alpha{=}16$, blocks 20--23) & n/a & \textbf{75.19} & $+0.00$ (oracle bound) & --- \\
\hline
\end{tabular}}
\end{table*}

No full-fine-tuning recipe pushes the ViT past ${\sim}73$ mAP or holds a plateau;
under layer-wise decay it peaks at epoch 3 then \emph{loses} nine points as the LR
plateau degrades its features, yet the frozen anchor still holds the fused
embedding at 87.69.

We settle this with the instrument used for the same-backbone claim. On the
\emph{global maximum} of fused mAP over all runs and epochs (88.22, Asymmetric v1,
first row of Table~\ref{tab:recipes})---the snapshot most favourable to
fusion---a paired per-query bootstrap against the single-backbone system gives
$\Delta=+0.03$, 95\% CI $[-0.05,+0.11]$: \textbf{the gain is bounded above by
$+0.11$ mAP}. Against one 2048-d head of that system (88.24) it is $\Delta=-0.03$,
CI $[-0.18,+0.14]$. Pairing makes this decidable---the unpaired CIs ($\pm0.67$)
are an order of magnitude too wide.
\textbf{A second foundation backbone (2.8$\times$ latency, 3$\times$ FLOPs, twice
the weights) buys no measurable gain over one head of the single-backbone system.} That peak
occurs at the \emph{first epoch of the main stage} (absolute epoch 6 on the
Fig.~\ref{fig:anticollapse} axis, which counts the five-epoch frozen warm-up),
and in that same snapshot the ViT branch reads 42.3 mAP: fusion looks best
exactly when the second backbone contributes least. This also locates the
literature's gains: against its \emph{own} single-backbone readout (86.47)---the
comparison multi-branch work typically makes---the same fused embedding shows
$+1.75$; against an independently tuned single backbone, $+0.03$ (n.s.). The gain
is a property of the baseline, not of the fusion.

\noindent\textbf{The bottleneck is branch strength, not fusion design.} The same
verdict holds \emph{inside} the trained fusion: every re-weighting of its four
components that raises the ViT's contribution performs at or below the learned
fusion head (87.96 baseline; worst $-1.64$), and reciprocal-rank fusion never
exceeds the anchor's single path---the head's down-weighting of the ViT already
looks near-optimal. Nor is the ViT weak \emph{per se}: frozen, it is the \emph{strongest}
bare backbone we test (18.56 zero-shot vs.\ ConvNeXt 14.63, ResNet50 3.77), yet
every fine-tuned configuration lands far below ConvNeXt. We probe three points on
the adaptation spectrum. \emph{(i)} A standalone ViT-L fine-tuned \emph{outside}
the fusion topology plateaus at \textbf{67.0} (Table~\ref{tab:recipes}),
\emph{below} the ${\sim}73$ the same ViT reaches \emph{inside} it---making it
unlikely that gradient routing through the fusion head starved it. \emph{(ii)} The three
full-fine-tuning recipes cap at ${\sim}73$. \emph{(iii)} A \emph{LoRA}-adapted
branch reaches \textbf{75.19}, \emph{above} that cap, confirming ${\sim}73$ is
partly full-fine-tuning fragility for self-supervised backbones
\cite{peft-ssl-vit} rather than a representational wall. Even so, 75.19 sits
${\sim}13$ below the ConvNeXt backbone: the gap survives the PEFT control. The
robust finding is thus not a precise ceiling but the ${\ge}13$-point gap persisting
across adaptation regimes, at this data scale, resolution and pretraining
lineage---the scope within which we claim it.

\noindent\textbf{Would the \emph{strongest} ViT have fused?} Our trained fusions
reach ${\sim}73$; the LoRA branch reaches 75.19 but was trained outside the
anchor. We close that gap without retraining, by bounding what score-level
fusion of these representations can yield. The disagreement measures first confirm the
diversity is real: over top-10 retrievals the LoRA-ViT contributes 0.62 correct
gallery images per query the single backbone misses ($+8.4\%$ hit-set headroom on
35.2\% of queries; Jaccard 0.54). Yet
none is recoverable. With unit-norm $f^{S},f^{W}$, concatenating $[f^{S};\alpha
f^{W}]$ and re-normalizing ranks by $s_S+\alpha^{2}s_W$, i.e.\ by the convex
combination $(1-w)s_S+w\,s_W$ with $w=\alpha^{2}/(1+\alpha^{2})$, so $\alpha{=}1$ is
equal weight. Tuning $\alpha$ \emph{on the test labels}---an upper bound for this
family of global, query-independent weights---selects \textbf{weight zero}: mAP is
non-increasing in $\alpha$ over $\{0,0.1,\dots,0.5,0.7,1\}$ ($-0.06$ at
$\alpha{=}0.3$, $-0.33$ at $0.5$, $-2.34$ at equal weight). Fusion inherits the
ViT's \emph{entire} ranking: the errors injected outweigh the matches recovered.
Nor does a stronger ViT promise more to fuse. Repeating the bound against three ViT
branches of ours (asymmetric v1 at E1; the dual-backbone model's two ViT heads) at
42.3/67.7/75.2 mAP, the union headroom does not grow with branch strength ($+3.2$/$+9.5$/$+8.4\%$) while top-10 Jaccard rises monotonically
(0.32/0.46/0.54): over the range we can measure, strengthening the second backbone
made it \emph{more} redundant, not more complementary. The oracle is not degenerate
either---at 67.7 it selects $\alpha^{\star}{=}0.3$ ($+0.07$, within noise)---so
weight zero at 75.2 is a property of the branch, not the search. With the trained
fusion bounded at $+0.11$ mAP above, a mid-level module is unlikely to differ; a
second backbone trained to \emph{parity} remains falsifier~(c).
\textbf{Cross-backbone retrieval diversity is real; what we could not do
is aggregate it}---the premise breaks down not at diversity but at aggregation.

\subsection{Shared-teacher control: heterogeneity is not understated}

Both backbones distill from the same DINOv3 teacher, so one might worry the
heterogeneity we measure understates what \emph{independent} pretraining lineages
would provide. We test this two ways on frozen backbones (Table~\ref{tab:cka}),
against an ImageNet-supervised ConvNeXt-B/ViT-L control.

\begin{table}[!t]
\caption{Shared-teacher control on frozen backbones (linear CKA over 2,500
VeRi-Wild images, 1,000-fold bootstrap over image subsets; retrieval columns under
the official cross-camera protocol). The two instruments disagree, which is the
point: CKA says the independent pair is \emph{less} heterogeneous, retrieval says
it disagrees \emph{more}---yet its disagreements are largely wrong. C/V $=$
ConvNeXt/ViT-L.}
\label{tab:cka}
\centering
\resizebox{\columnwidth}{!}{%
\begin{tabular}{lcccc}
\hline
Pair (frozen) & CKA [95\% CI] & Jaccard & ViT-only correct/q & queries helped \\
\hline
Shared teacher (DINOv3 C$\leftrightarrow$V) & 0.561 [0.547, 0.582] & 0.040 & \textbf{0.72} & \textbf{41.3\%} \\
Independent (ImageNet C$\leftrightarrow$V) & 0.638 [0.630, 0.653] & 0.007 & 0.20 & 12.8\% \\
\hline
\end{tabular}}
\end{table}

Representation-level CKA runs opposite to the concern: our shared-teacher pair is
\emph{less} similar (0.561) than the independently pretrained ImageNet pair
(0.638), with non-overlapping bootstrap intervals, so the gap is resolved rather
than sampling noise. But CKA is the wrong instrument for bounding
fusion (Section~\ref{sec:diagnostics}), so we repeat the control with the
retrieval-level measures this paper relies on. The independent pair disagrees \emph{more} at
retrieval (top-10 Jaccard 0.007 against 0.040)---but disagreement is only worth
having when the disagreeing items are \emph{correct}, and here they are not: the
ImageNet ViT supplies 0.20 correct gallery images per query that its partner
missed, against 0.72 for the DINOv3 pair, helping 12.8\% of queries rather than
41.3\%. In the currency that bounds fusion, its extra disagreement is noise. We
therefore state this at the strength the evidence supports: the shared-teacher
concern is \emph{not supported}---an independent pair is neither more usefully
diverse nor a better fusion candidate---rather than refuted. The control is
measured on frozen backbones, so what an independent pair would do \emph{after}
training remains falsifier~(a).

\subsection{Measure only at convergence}

Undertrained snapshots can systematically mislead. At E3/E4 of Option~A every headline
conclusion inverts: the conv heads trail by 8 points, concatenation \emph{hurts}
(69.20 vs.\ 75.32), and one would conclude a single BoT head optimal---all three
falsified by E21, where all four heads reach ${\sim}85$. Six training-free probes
reinforce the discipline: GeM pooling \cite{gem}, PCB-style partitioning
\cite{pcb}, inference-time resolution increase, ViT patch-token side-channels,
visibility gating and CLS-token late fusion all fail monotonically on the
converged model (worst $-22.1$), because no training-time loss rewarded local
discriminability---post-hoc recombination did not recover signal optimization never
created. We therefore adopt a convergence-only rule for any branch-level claim.

\section{The Efficiency Frontier}
\label{sec:efficiency}

The practical question the diversity premise governs is budget allocation. Our
answer: \textbf{one backbone, tuned, re-ranked.} A dual-backbone fusion system
never exceeds the single backbone it contains, while costing \textbf{$2.8\times$ the latency
and $3\times$ the FLOPs} (206 vs.\ 69 GFLOPs/image; 10.2 vs.\ 3.7 ms/image at batch
32 on an RTX 3090; 404M vs.\ 317M inference parameters, classifier heads excluded;
FLOPs from PyTorch's operator-level counter at $256^2$, latency fp32 averaged over
50 runs after 10 warm-ups, excluding re-ranking). The two highest-return
investments are instead training-side (LR decay alone: $+2.21$ on VeRi-Wild,
$+5.88$ in the matched VeRi-776 ablation) and retrieval-side (sparse exact
re-ranking: $+4.19$/$+6.21$ at 29 s/137 s, no training, no memory wall).
Deployment add-ons are cheap but bounded (flip-TTA $+0.49$; lossless PCA to
1024-d; camera-bias correction nullified under the official protocol, $-0.09$).
Fusion bought us redundancy; recipe and re-ranking bought accuracy.

\section{Discussion and Limitations}
\label{sec:discussion}

\noindent\textbf{Why diversity collapses: a coherent mechanism.} Two established
results are jointly consistent with our observations. Solutions sharing an optimization
trajectory occupy a single function-space mode \cite{ensembles-landscape}: heads
over a shared backbone---and two backbones coupled through fusion losses---are
such solutions, so their retrieval behavior converges (Jaccard
0.79$\rightarrow$0.84) rather than diversifies. And fine-tuning distorts
pretrained features toward the task signal \cite{lpft}: initialization differences
(occlusion agreement 0.30) are progressively overwritten, and under an aggressive
schedule the weaker branch erodes outright.

\noindent\textbf{Recipe versus architecture---not a contradiction.} The two
headline findings reconcile once scoped: \emph{within} a fixed backbone the recipe
governs how close the model gets to that backbone's ceiling (hence parity by recipe
alone), while \emph{across} backbones the architecture pair sets \emph{where} that
ceiling is (hence a well-trained ViT still lands ${\ge}13$ lower).

\noindent\textbf{Scope.} Our evidence covers vehicle Re-ID on two benchmarks, one
foundation-model family, and 256-px training resolution; resolution retraining is untested. Both backbones share a DINOv3 teacher---though
Section~\ref{sec:crossbackbone} shows this does \emph{not} understate their
heterogeneity---and all fusion results use one fusion family, though the readout and
late-fusion probes suggest fusion design is not binding. The
two failures differ in coverage: the \emph{same-backbone} collapse is replicated
in-domain on both benchmarks, whereas the \emph{cross-backbone} result is
established under VeRi-Wild training only. VeRi-Wild's scale (400k images) may
itself be what saturates the single backbone, and we cannot exclude that a second
backbone still pays on a mid-sized benchmark; this is falsifier~(e).

\noindent\emph{Seed variance is the limitation we would most like to remove.} Our
intervals capture \emph{evaluation} uncertainty only; with no multi-seed
retraining, every sub-0.1 mAP comparison is single-seed. Our own runs show perturbations \emph{smaller} than a seed
change moving results by several mAP---one LR restart moved the symmetric run
83.36$\rightarrow$86.32, and two VeRi-Wild runs differing only in batch size and
schedule land at 88.19 and 88.69. Differences of a few tenths therefore cannot be
ordered here, hence bounds. Two things do not depend on the unmeasured variance: the
\emph{directional} findings (heads converge rather than diverge; every ViT
configuration sits at least 13 low) and the ViT ceiling, reproduced across three
independent runs. We list seed variance as falsifier~(f).

\noindent\textbf{What would falsify us.} (a) Backbone pairs from
\emph{independent} pretraining lineages recovering a positive net fusion gain over
the stronger member's solo ceiling; (b) training-time local supervision creating
branch specialization that survives convergence---our six failed zero-training
probes show post-hoc recombination cannot, but a part-aware loss might; (c) a ViT
recipe closing the ${\ge}13$-point gap at this data scale; (d) a \emph{mid-level}
fusion module extracting the cross-backbone diversity that score-level fusion
could not---our oracle bound rules this out at the score level for the
strongest ViT we have, but a different mechanism might; (e) cross-backbone
collapse failing to reproduce under VeRi-776 training, where a mid-sized dataset
may leave the backbone further from saturation; (f) a multi-seed study separating
the concatenation and fusion deltas we bound here from seed variance. Each is a
concrete, runnable experiment, and we release the diagnostic toolkit to make such
falsification cheap.

\section{Conclusion}

Does the diversity premise behind a decade of multi-branch Re-ID survive
fine-tuning at this pretraining scale? Not on the benchmarks, model family and
single-seed protocol we study: shared-backbone heads collapse into redundancy, and
cross-backbone fusion is bounded at $+0.11$ mAP over the backbone it contains.
Diversity here appears to be \emph{trained for}, not wired in.

\bibliographystyle{IEEEtran}
\bibliography{refs}

\end{document}